# Recognition of Non-Compound Handwritten Devnagari Characters using a Combination of MLP and Minimum Edit Distance

**Sandhya Arora** sandhyabhagat@yahoo.com
*Assistant Professor, Department of CSE & IT*
*Meghnad Saha Institute of Technology*
*Kolkata-107, India*

**Debotosh Bhattacharjee**
*Department of Computer Science and Engg.*
*Jadavpur University*
*Kolkata-107, India*

**Mita Nasipuri**
*Department of Computer Science and Engg.*
*Jadavpur University*
*Kolkata-107, India*

**D. K. Basu**
*Department of Computer Science and Engg.*
*Jadavpur University*
*Kolkata-107, India*

**M. Kundu**
*Department of Computer Science and Engg.*
*Jadavpur University*
*Kolkata-107, India*

**Abstract**

This paper deals with a new method for recognition of offline Handwritten non-compound Devnagari Characters in two stages. It uses two well known and established pattern recognition techniques: one using neural networks and the other one using minimum edit distance. Each of these techniques is applied on different sets of characters for recognition. In the first stage, two sets of features are computed and two classifiers are applied to get higher recognition accuracy. Two MLP's are used separately to recognize the characters. For one of the MLP's the characters are represented with their shadow features and for the other chain code histogram feature is used. The decision of both MLP's is combined using weighted majority scheme. Top three results produced by combined MLP's in the first stage are used to calculate the relative difference values. In the second stage, based on these relative differences character set is divided into two. First set consists of the characters with distinct shapes and second set consists of confused characters, which appear very similar in shapes. Characters of distinct shapes of first set are classified using MLP. Confused characters in second set are classified using minimum edit distance method. Method of minimum edit distance makes use of corner detected in a character image using modified Harris corner detection technique. Experiment on this method is carried out on a database of 7154 samples. The overall recognition is found to be 90.74%.

## 1. INTRODUCTION

Optical Character Recognition (OCR) is the most crucial part of Electronic Document Analysis Systems. The solution lies in the intersection of the fields of pattern Recognition, image and natural language processing. Although there has been a tremendous research effort, the state of the art in the OCR has only reached the point of partial use in recent years. Nowadays, clearly printed texts in documents with simple layouts can be recognized reliably by off-the-shelf OCR software. There is only limited success in handwriting recognition, particularly for isolated and neatly hand-printed characters and words for limited vocabulary. However, in spite of the intensive effort of more than thirty years, the recognition of free style handwriting continues to remain in the research arena.

An OCR has variety of commercial and practical applications in processing bank cheques, government records, credit card imprints and postal code reading, reading commercial forms, manuscripts and their archival etc. Such a system facilitates a key board less user computer interaction also the text which is either printed or handwritten can be directly transferred to the machine. An elaborate list of OCR applications has been presented by Govindan[1].

Historically, Devnagari is the script used by Sanskrit, Hindi, Marathi and Nepali. Hindi is the world's third most commonly used language after Chinese and English. Thus research on Devnagari script is gaining importance because of their large market potential. With the explosion of information technology there has been a dramatic increase of research in this field since the beginning of 1980.

OCR work on printed Devnagari Script started in early 1970's. Sinha and Mahabala[2] presented a syntactic pattern analysis system with an embedded picture language for the recognition of handwritten and machine printed Devnagari characters. Veena described Devnagari OCR in her doctoral Thesis [3]. Performance of 93% accuracy at character level is reported after post processing. Pal and Chaudhuri [4] reported a complete OCR system for printed Devnagari with 96% accuracy. Hanmandlu and Murthy [5,6] proposed a Fuzzy model based recognition of handwritten Devnagari numerals and characters and they obtained 92.67% accuracy for Handwritten Devnagari numerals and 90.65% accuracy for Handwritten Devnagari characters. Sinha et al [2,7] have reported various aspects of Devnagari script recognition. Bajaj et al [8] employed three different kinds of features namely, density features, moment features and descriptive component features for classification of Devnagari Numerals. They proposed multi-classifier connectionist architecture for increasing the recognition reliability and they obtained 89.6% accuracy for handwritten Devnagari numerals. Kumar and Singh [9] proposed a Zernike moment feature based approach for Devnagari handwritten character recognition. They used an artificial neural network for classification. Sethi et. al. [10,11] has described Devnagari numeral recognition based on structural approach. The primitive used are horizontal line segment, vertical line segment, right slant and left slant. A decision tree is employed to perform analysis based on presence/absence of these primitives and their interconnection. A similar strategy was applied to constrained hand printed Devnagari character. Bansal et. al. [12], have used translation and scaling invariant moments and structural description of a character and reported accuracy of 93% at character level of printed Devnagari characters. Bhattacharya et al [13] proposed a Multi-layer perceptron (MLP) neural network based classification approach for the recognition of Devnagari handwritten numerals and obtained 91.28% results. They considered a multi- resolution features based on wavelet transform in their proposed system. N. Sharma and U. Pal [14,15,16] proposed a directional chain code features based quadratic classifier and obtained 80.36% accuracy for handwritten Devnagari characters and 98.86% accuracy for handwritten Devnagari numerals. Few more work[17,18,19] is going on for handwritten devnagari characters. In our previous work [20] we proposed a MLP based classifier designed with three different features namely: Intersection, Shadow, Chain code histogram. The recognition accuracy 69.37% achieved by considering top 1 choices results on 4900 samples. In this paper, we purpose a system based on MLP and minimum edit distance for the recognition of offline Handwritten Devnagari character recognition.

While a large amount of literature is available for recognition of English script, relatively less work has been reported for the recognition of Indian languages. Main reason for this slow development could be attributed to the complexity in the shapes of Indian scripts, and also the large set of different patterns that exists in these languages, as opposed to English.

Most of the work reported above [2,4,12] were on printed Devnagari characters. For handwritten Devnagari character recognition system [3,5,6,9] , accuracy reported is not high and dataset used are not large. We worked on 7154 samples. As no standard database is available for handwritten Devnagari characters, we created some samples of our own and some we collected from ISI, Kolkata.

Rest of the paper is organized as follows. In section 2, peculiarities of Devnagari Script are discussed. Overall approach used, is discussed in section 3. Feature extraction techniques are reported in section 4. Section 5, deals with the classifiers used for the recognition purpose. The experimental results are discussed in section 6.

## 2. PECULIARITIES OF DEVNAGARI SCRIPT

Devnagari script is different from Roman script in several ways. This script has two-dimensional compositions of symbols: core characters in the middle strip, optional modifiers above and/or below core characters. Two characters may be in shadow of each other. While line segments (strokes) are the predominant features for English, most of the characters in Devnagari script is formed by curves, holes, and also strokes. In Devnagari language scripts, the concept of upper-case, the lower-case characters is absent. However the alphabet itself contains more number of symbols than that of English.

Devnagari script has around 13 vowels and 36 consonants resulting in a total of 49 basic characters as shown in Figure 1a. Vowels occur either in isolation or in combination with consonants. Apart from vowels and consonants characters called basic characters, there are compound characters in Devnagari script alphabet system, which are formed by combining two or more basic characters (shown in Figure 1b). The shape of compound character is usually more complex than the constituent basic characters. Coupled to this in Devnagari script there is a practice of having more than twelve forms each for 36 consonants , giving rise to modified shapes which, depending on whether the vowel modifier is placed to the left, right, top or bottom of the consonants as shown in Figure 1c. They are called modified characters. The net result is that there are several thousand different shapes or patterns in the script, some of them are almost similar in shapes. Even with the basic character same problem about their shapes exists. Some basic characters have distinct shapes (Figure 1a) and can be identified with certainty. Some groups of basic characters have almost similar shapes (Figure 1d) causing confusion and need special attention in recognizing them. The most of the confusing pair of Devnagari characters are from the Figure specified in 1d.

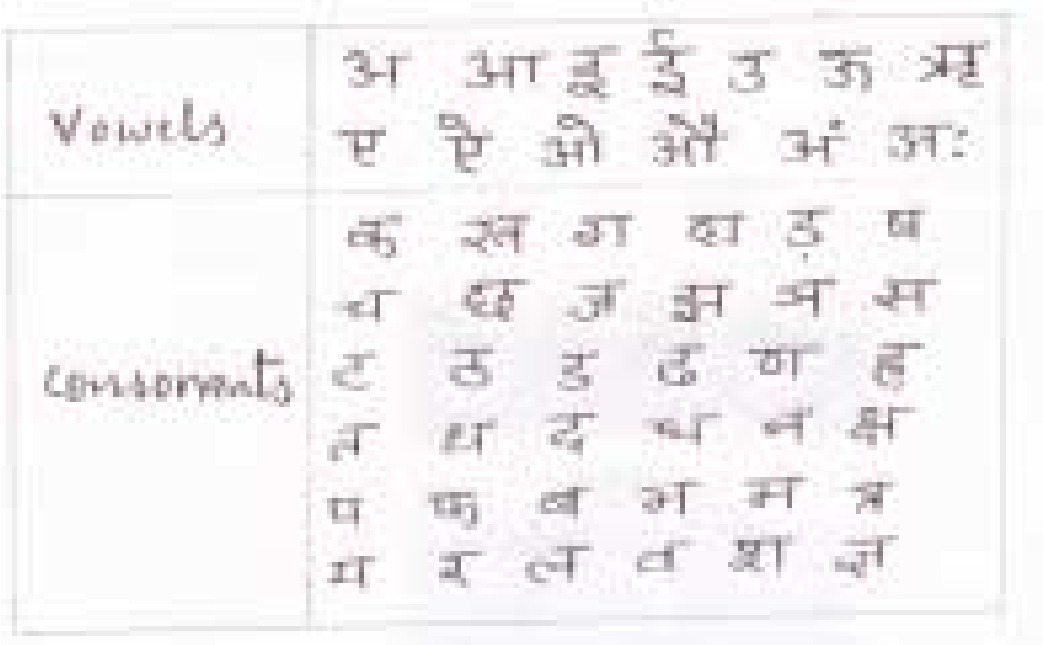

(a)

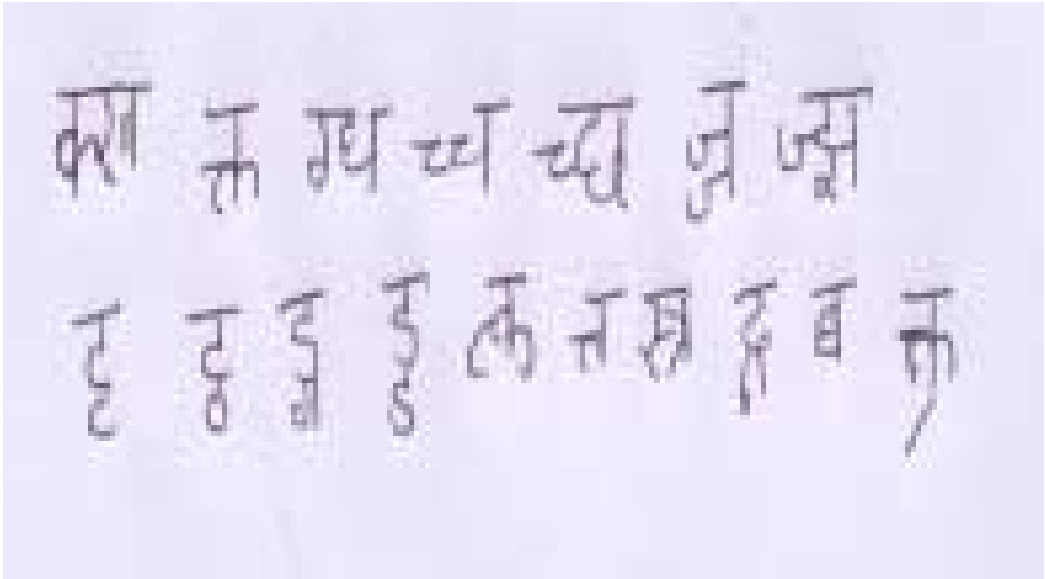

(b)

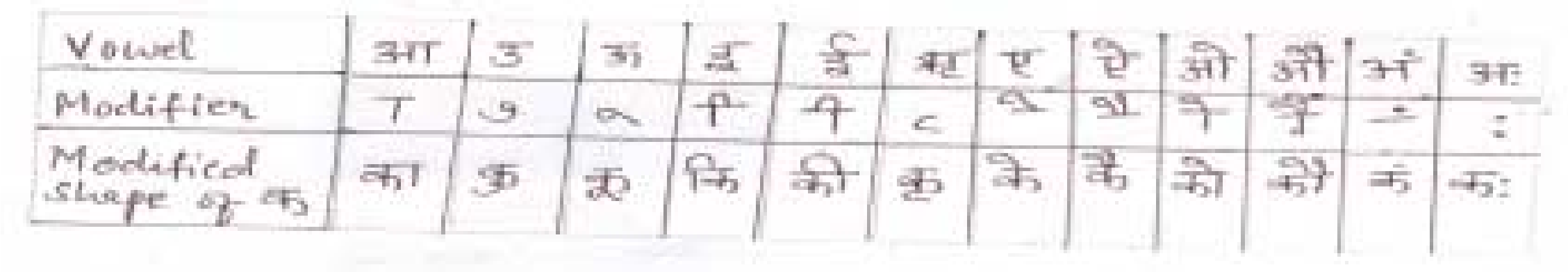

( c)

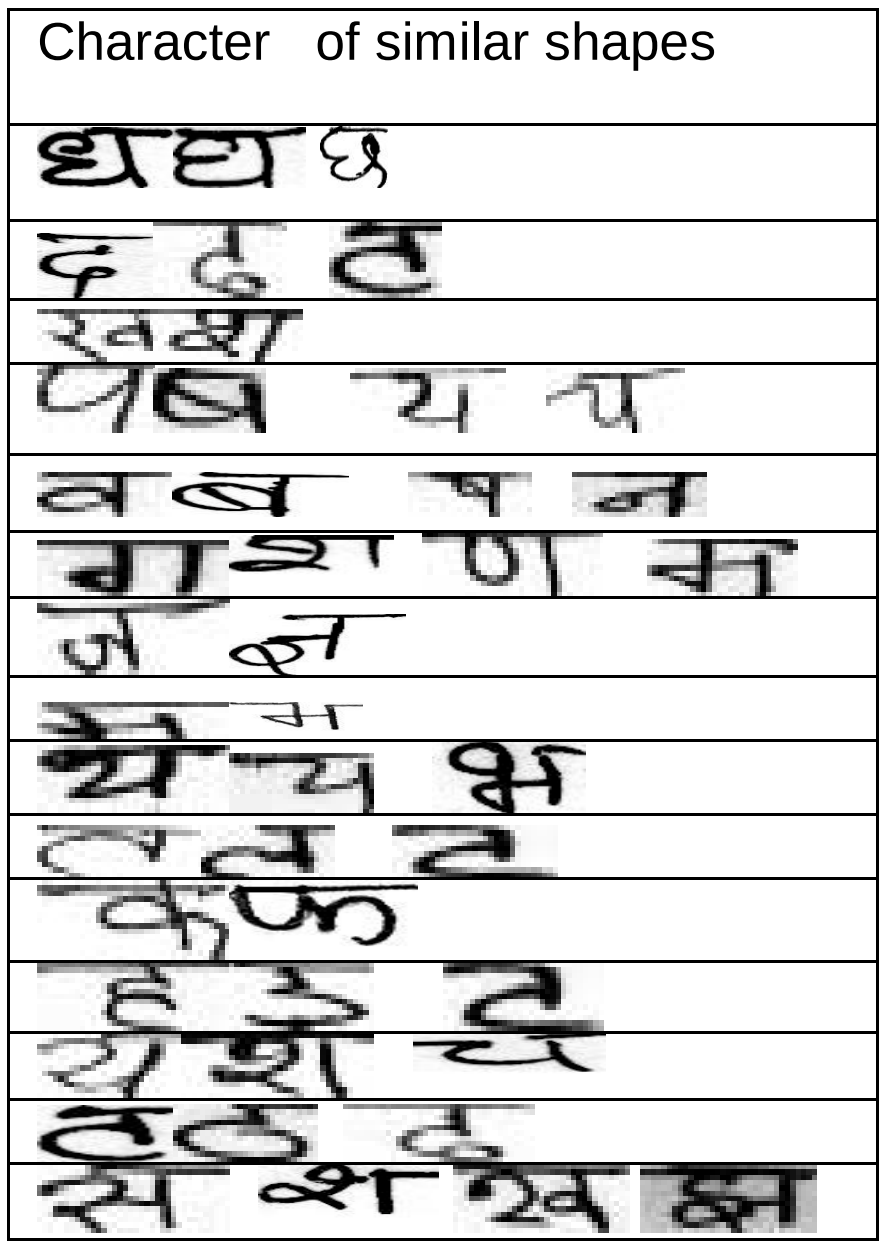

(d)

**FIGURE 1**. Samples of handwritten devnagari a)Vowels and Consonants b) some compound characters c) Modifiers with their corresponding vowel and a sample character image of “ka” modified with modifier d) Confusing characters

## 3. OVERALL APPROACH

Scheme of our proposed method is shown in Figure 2. We perform scaling of character bitmap and after that we extract two different features. First, 24 shadow features are extracted from eight octants of the scaled binarized character image. Second, 200 chain code histogram features are obtained by first detecting the contour points of original scaled binarized character image, and dividing the contour image into 25 segments. For each segment chain code histogram features are obtained. Here, the character recognition is done in two stages. In the first stage, two MLP’s are designed using these two different feature sets. Outputs of individual MLP classifiers [20] are combined using weighted majority scheme and the character classes corresponding to top three values are considered. A relative difference measure is computed from these top three values. If this measure is greater than some threshold value, we infer that the top choice determines the class of the sample character with certainty. On the other hand, if the relative difference measure is less than or equal to the threshold value, we infer that the sample character belongs to a group of confusing character identified by top three choices. In the second stage , the true class of the sample character belonging to a confusing group are identified by applying minimum edit distance method, on detected corners of the sample character using a modified form of Harris corner detector.

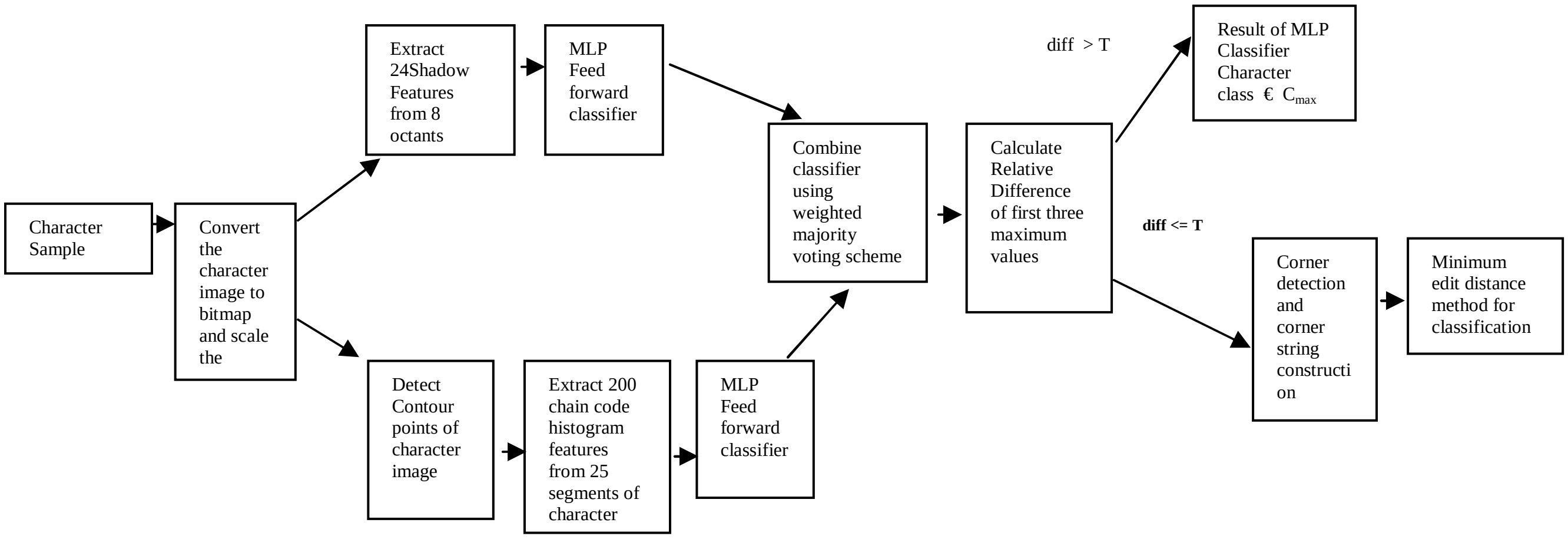

**FIGURE 2**. Overall scheme of proposed Technique

## 4. FEATURE EXTRACTION

In the following section we give a brief description of the two feature sets used in our proposed system. Shadow features are extracted from scaled binarized character image. Chain code histogram features are extracted by chain coding the contour points of the scaled character binarized image.

### 4.1 Shadow Features of a character image

Shadow is basically the length of the projection on the sides as shown in Figure 3. For computing shadow features [21], the rectangular boundary enclosing the character image is divided into eight octants. For each octant shadows or projections of character segment on three sides of the octant dividing triangles are computed so, a total of 24 shadow features are obtained. Each of these features is divided by the length of the corresponding side of the triangle to get a normalized value.

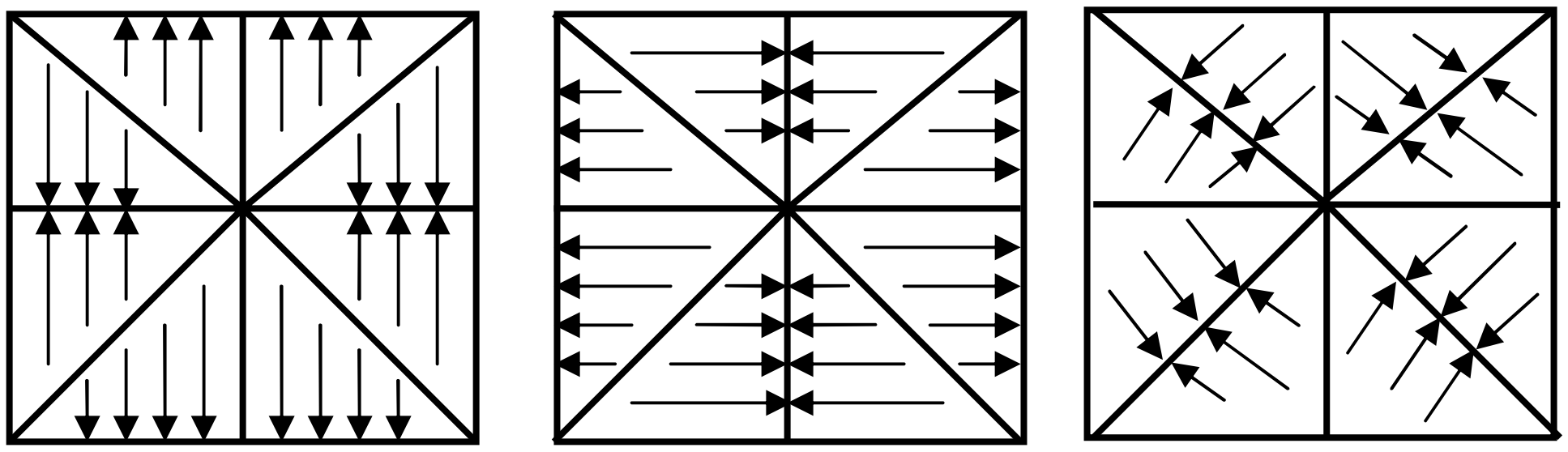

**FIGURE 3.** Shadow features

### 4.2 Chain Code Histogram of Character Contour

Chain code provides the direction of the next pixel in the image. Given a scaled binary image, we first find the contour points of the character image. We consider a 3 × 3 window surrounded by the object points of the image. If any of the 4-connected neighbor points is a background point then the object point (P), as shown in Figure 4 is considered as contour point.

| | | |
|---|---|---|
| | X | |
| X | P | X |
| | X | |

**FIGURE 4**. Contour point detection

The contour following procedure is used to trace the contour and a contour representation called “chain coding” as proposed by Freeman [22], shown in figure 5a. Each pixel of the contour is assigned a different code that indicates the direction of the next pixel that belongs to the contour in some given direction. Chain code provides the points in relative position to one another, independent of the coordinate system. In this methodology of using a chain coding of connecting neighboring contour pixels, the points and the outline coding are captured. Contour following procedure may proceed in clockwise or in counter clockwise direction. Here, we have chosen to proceed in a clockwise direction.

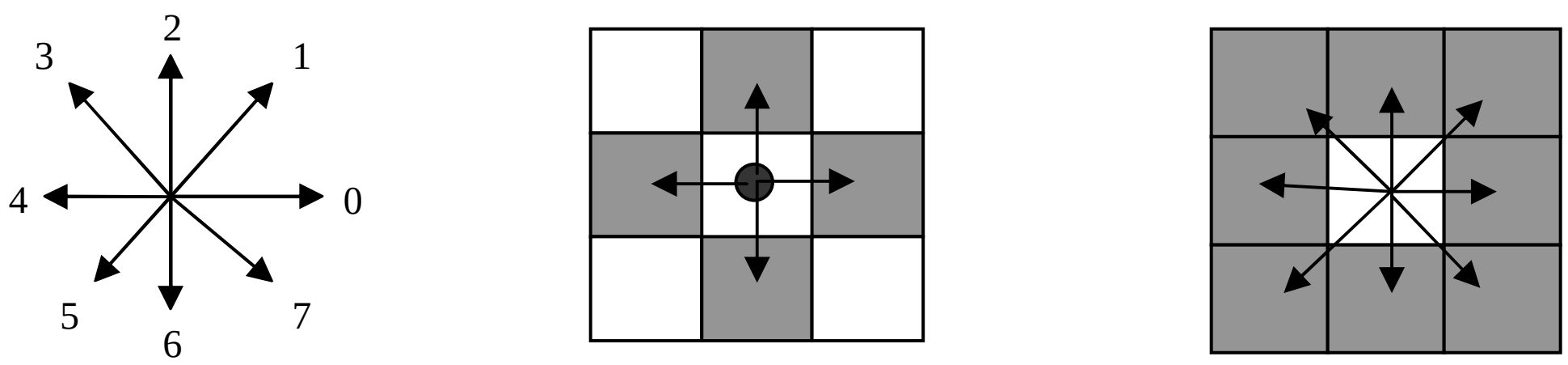

(a) (b) (c)

**FIGURE 5**. Chain Coding: (a) direction of connectivity, (b) 4-connectivity, (c) 8-connectivity. Generate the chain code by detecting the direction of the next-in-line pixel

The chain code for the character contour will yield a smooth, unbroken curve as it grows along the perimeter of the character and completely encompasses the character. When there is multiple connectivity in the character, then there can be multiple chain codes to represent the contour of the character. We chose to move with minimum chain code number first.

We divide the contour image in 5 × 5 blocks. In each of these blocks, the frequency of the direction code is computed and a histogram of chain code is prepared for each block. Thus for 5 × 5 blocks we get 5 × 5 × 8 = 200 features for recognition.

## 5. CHARACTER RECOGNITION

Devnagari character recognition is done in two stages using Multilayer Perceptron (MLP) and method of minimum edit distance, each applied at different stage. We divided the character set in two set using relative difference value discussed in section 6.1. First set consists of characters with certainty and second set consists of confused characters. Characters of first set are classified using MLP discussed in section 5.1 and characters of second set are classified using minimum edit distance method applied on corner detected character image, is discussed in section 5.2. We rejected some samples in MLP classifier and the corner detection method with minimum edit distance is applied to these confused rejected samples to increase the accuracy.

### 5.1 MLP Classifier

We designed different MLP with 3 layers including one hidden layer for two different feature sets consisting of 24 shadow features and 200 chain code histogram features. The experimental results obtained while using these features for recognition of handwritten Devnagari characters is presented in section 6. At this stage all characters are non-compound, single characters so no segmentation is required.

Each MLP is trained with Backpropagation learning algorithm with momentum [13]. It minimizes the sum of squared errors for the training samples by conducting a gradient descent search in the weight space. As activation function we used sigmoid function. Learning rate and momentum term are set to 0.8 and 0.7 respectively. As activation function we used the sigmoid function. Numbers of neurons in input layer of MLPs are 24 or 200, for shadow features and chain code histogram features respectively. Number of neurons in Hidden layer is not fixed, we experimented on the values between 20-70 to get optimal result and finally it was set to 30 and 70 for shadow features and chain code histogram features respectively. The output layer contained one node for each class., so the number of neurons in output layer is 49.

### 5.1.1 Combining Multiple Classifiers

The ultimate goal of designing pattern recognition system is to achieve the best possible classification performance. This objective traditionally led to the development of different classification scheme for any pattern recognition problem to be solved. Classifiers producing crisp, single class labels (SCL) provide the least amount of useful information for the combination process. However, they are still well performing classifiers and the sets of patterns misclassified by the different classifiers does not necessarily overlap. This suggested that different classifier designs potentially offered complementary information about the pattern to be classified which could be harnessed to improve the performance of the selected classifier. So instead of relying on a single decision making scheme we can combine classifiers.

Voting strategies can be applied to a multiple classifier system assuming that each classifier gives a single class label as an output. There are a number of approaches to combination of such uncertain information units in order to obtain the best final decision [21]. We applied the voting definition. For convenience let the output of the classifiers form the decision vector d defined as d $=[d_1, d_2, \ldots, d_m]^T$ where $d_i \in \{c_1, c_2, \ldots, c_m, r\}$, $c_i$ denotes the label of the i-th class and r the rejection of assigning the input sample to any class. Let binary characteristic function be defined as follows:

$$B_j(c_i)=\begin{cases}1 & \text{if } d_j = c_i\\ 0 & \text{if } d_j \neq c_i\end{cases}$$

Then the general voting routine can be defined as:

$$E(d)=\begin{cases}c_i & \text{if } \forall_{t\in\{1....m\}} \sum_{j=1}^{n} B_j(c_t) \le \sum_{j=1}^{n} B_j(c_t) \ge \alpha.m + k(d)\\ r & \text{otherwise}\end{cases}$$

Where α is a parameter and k(d) is a function that provides additional voting constraints. The most conservative voting rule is given if k(d) =0 and α =1 ,meaning that the class is chosen when all classifiers produce the same output. This rule can be liberalized by lowering the parameter α. Function k(d) is usually interpreted as a level of objection to the most often selected class and refers mainly to the score of the second ranked class. This option allows adjusting the level of collision that is still acceptable for giving correct decision.

$$\alpha = \frac{p_k}{\sum_{k=1}^{2} p_k}$$

In our present work, we have used two MLP classifiers with recognition performances

$p_1$=73.33% (Success rate of MLP1 using chain code histogram feature)
$p_2$= 68.10% (Success rate of MLP2 using shadow feature)

### 5.2 Confused characters classification

For classifying characters of similar shapes we took different approach. We detected corners in character image using modified form of Harris corner detector [23], discussed in section 5.2.1 and 5.2.2. After detecting corners we divided the character image in 25 segments and in each segment we counted the number of corners. On this corner detected string we applied minimum edit distance discussed in section 5.2.3

#### 5.2.1 Corner Detection Algorithm

Corner Detector can be considered interest point corner detector as they assign a measure of cornerness to all pixels in an image. The brute force method of comparing every pixel in the two images is computationally prohibitive. Intuitively one can relate two images by matching only locations in the image that are in some way interesting. Such points are referred to as interest points and are located using an interest point detector. Finding a relationship between images is then performed using only these points. This drastically reduces the required computation time. Corner points are interesting as they are formed from two or more edges and edges usually define the boundary between two different objects or parts of the same object.

Many Corner Detectors[23] are available but we chose Harris/Plessey corner detector with some modification. This corner detector is computationally demanding, but directly addresses many of the limitations of the other corner detectors.

Algorithm for detecting corners using Harris Corner detector in all confused characters is as follows:-

1. For each pixel (x, y) in the image calculate the autocorrelation matrix M:-

   $$M = \begin{bmatrix} A & C \\ C & B \end{bmatrix}$$

   Where $A = \left(\frac{\partial I}{\partial x}\right)^2 \otimes w$

   $$B = \left(\frac{\partial I}{\partial y}\right)^2 \otimes w$$

   $$C = \left(\frac{\partial I}{\partial x}\frac{\partial I}{\partial y}\right)^2 \otimes w$$

   $\otimes$ is the convolution operator , W is the Gaussian window of size 5

   $\left(\frac{\partial I}{\partial x}\right)$, $\left(\frac{\partial I}{\partial y}\right)$, $\left(\frac{\partial I}{\partial x}\frac{\partial I}{\partial y}\right)$ are horizontal, vertical and diagonal intensity variations respectively.

2. Construct the cornerness map by calculating the cornerness measure C(x, y) for each pixel (x, y):

   ***$C(x,y) = det(M)-k(trace(M))^2$***
   ***$det(M) = \lambda_1 \lambda_2 = AB-C^2$***
   ***$trace(M) = \lambda_1 + \lambda_2 = A + B$***
   ***$k$ = constant***
   $\lambda_1$ and $\lambda_2$ eigenvalues of M

3. Threshold the interest map by setting all C(x, y) below a threshold T to zero.
4. Perform non-maximal suppression to find local maxima.

The basic idea behind detecting corners of an image in this algorithm is to estimate the measurement of local autocorrelation so intensity variation is measured in different directions for that purpose. Intensity variation calculation for Harris operator is approximated using the gradient of the image so the intensity variation in horizontal, vertical and diagonal direction can be written as a function of the gradient of the image. Here A, B, C in step 1 is defined as:-

A= Weighted horizontal intensity variation
B= Weighted vertical intensity variation
C= Weighted diagonal intensity variation

Here the intensity variation is convolved with the Gaussian window. Gaussian window is a circular window that puts more weight on measurement made closer to the centre of the window. This is desirable so that the Euclidean distance from the centre pixel to the edge is same in all directions. This improves the estimate of the local intensity variation.

Weighted horizontal intensity variation A is calculated by convolving below specified window in figure 6 with Gaussian window of size 5. Vertical Intensity Variation B is calculated by convolving below specified window in figure 7 with Gaussian window of size 5. Diagonal Intensity Variation in upward direction C is calculated by convolving below specified window in figure 8 with Gaussian window of size 5.

| | | | |
|---|---|---|---|
| A1 | A2<br>B1 | A3<br>B2 | B3 |
| A4 | A5<br>B4 | A6<br>B5 | B6 |
| A7 | A8<br>B7 | A9<br>B8 | B9 |

**FIGURE 6.** Horizontal intensity variation window(A)

| | | |
|---|---|---|
| B1 | B2 | B3 |
| A1<br>B4 | A2<br>B5 | A3<br>B6 |
| A4<br>B7 | A5<br>B8 | A6<br>B9 |
| A7 | A8 | A9 |

**FIGURE 7**. Vertical intensity variation window(B)

| | | | |
|---|---|---|---|
| | B1 | B2 | B3 |
| A1 | A2<br>B4 | A3<br>B5 | B6 |
| A4 | A5<br>B7 | A6<br>B8 | B9 |
| A7 | A8 | A9 | |

**FIGURE 8.** Diagonal upward intensity variation window(C)

| | | | | |
|---|---|---|---|---|
| W1<br>.004 | W2<br>.015 | W3<br>.026 | W4<br>.015 | W5<br>.004 |
| W6<br>.015 | W7<br>.059 | W8<br>.095 | W9<br>.059 | W10<br>.015 |
| W11<br>.026 | W12<br>.095 | W13<br>.15 | W14<br>.095 | W15<br>.026 |
| W16<br>.015 | W17<br>.059 | W18<br>.095 | W19<br>.059 | W20<br>.015 |
| W21<br>.004 | W22<br>.015 | W23<br>.026 | W24<br>.015 | W25<br>.004 |

**FIGURE 9**. Gaussian Window of size 5

### 5.2.2 Modification to Harris Corner Detection Algorithm

We modified this algorithm because it just considers weighted intensity variations in three directions i.e. in horizontal, vertical and diagonal upward direction. We took one more measure of C i.e. weighted diagonal intensity variation in downward direction D, because previous A, B, C measures alone does not detect corners properly for Devnagari characters. Now the Diagonal Intensity variation in downward direction D is calculated by convolving below specified window in figure 10 with Gaussian window of size 5.

| | | | |
|---|---|---|---|
| | A1 | A2 | A3 |
| B1 | B2<br>A4 | B3<br>A5 | A6 |
| B4 | B5<br>A7 | B6<br>A8 | A9 |
| B7 | B8 | B9 | |

**FIGURE 10** Diagonal downward intensity variation window (D)

So the autocorrelation matrix M in step 1 is modified as

$$M = \begin{bmatrix} A & C+D \\ C+D & B \end{bmatrix}$$

Steps 2, 3 and 4 are performed as specified above. All non-zero points remaining in the corner ness map are corners. We used k=0.04 in our algorithms.

**5.2.3 Method of Minimum edit Distance**

After detecting the corners in the character image we segmented it into 25, and in each segment we counted number of corner points. So for each character we got a corner string of 25 which is utilized for calculating the distance among the characters. Minimum edit distance method [17] is used for this purpose. Distance is a measure of similarity between two strings which is referred as source string and target string. The distance is the number of deletions, insertions or substitutions required to transform s into t. The basic idea behind calculating the distance D(i,j) between two corner string s1[1…i] and s2[1….j] is as follows:-

D(i,0)=i

$$D(i, j) = \text{minimum}\begin{pmatrix} D(i-1, j)+1, \\ D(i, j-1)+1, \\ D(i-1, j-1)+t(i, j) \end{pmatrix}$$

$$t(i, j) = \begin{cases} 0 & \text{if } s1(i) = s2(j) \\ 1 & \text{if } s1(i) != s2(j) \end{cases}$$

## 6. RESULTS

The experiment evaluation of the above technique was carried out on 7154 isolated samples of Devnagari basic characters (vowels as well as consonants) out of which 4900 samples were collected from ISI, Kolkata [24] and rest 2254 samples were collected from different people in our organization. A total of 65% characters are used for the training and rest is used for testing purpose. We have used 3-fold cross validation schema for recognition result computation. Data set is divided into 3 subsets and testing is done on each subset using the rest two subsets for learning. The recognition rates for all the test subsets are averaged to calculate recognition accuracy.

### 6.1 Recognition Results

The recognition accuracy obtained from our above discussed classifiers separately is shown in Table 1. The overall global recognition accuracy of our system using combined MLP is 76.67% when zero percent confusion was considered. 93.27% accuracy was obtained when we considered top 5 choices of the recognition result and with zero percent confusion.

| Classifier | Accuracy |
|---|---|
| Combined MLP (top 1 choices) | 76.67% |
| Combined MLP (top 5 choices) | 93.27% |
| Minimum Edit distance classifier | 85% |

**TABLE 1.** Individual accuracy of MLP and minimum edit distance method

### 6.2 Relative Difference versus Confused Characters

Confused character samples are separated out based on the relative difference value. Relative difference measure used is computed as

Diff = ( 2* max2-max1-max3 ) / ( 2*max2 )

Where max1, max2, max3 are top3 values of combined MLP classifier. If this relative difference is greater than some threshold value, we infer that the top choice determines the class of the sample character with certainty. On the other hand, if the relative difference measure is less than or equal to the threshold value, we infer that the sample character belongs to a group of confusing character identified by top three choices.

Considering the lowest relative difference value 57.3% characters were separated in characters with certainty set which have distinct shapes and rest 42.7% characters were in confused character set which are of very similar shapes. For classifying 57.3% characters MLP classifier was used which gave 97.27% accuracy and for 42.7% characters minimum edit distance classifier was used which gave 82% accuracy. The combined accuracy we achieved is 90.74%.

### 6.3 Confused Character Pair

From experiment we noticed that mainly the error occurred because of the similar shaped characters. Since the shape of the handwritten characters in this pair is very similar, most of the pair is misclassified. Some pair of Devnagari Characters which forms the main confusion pairs of characters is listed in table 2. Maximum error occurred between and , and , and , and , and , and , and , and , and , and , and , and , and , and , and , and pairs.

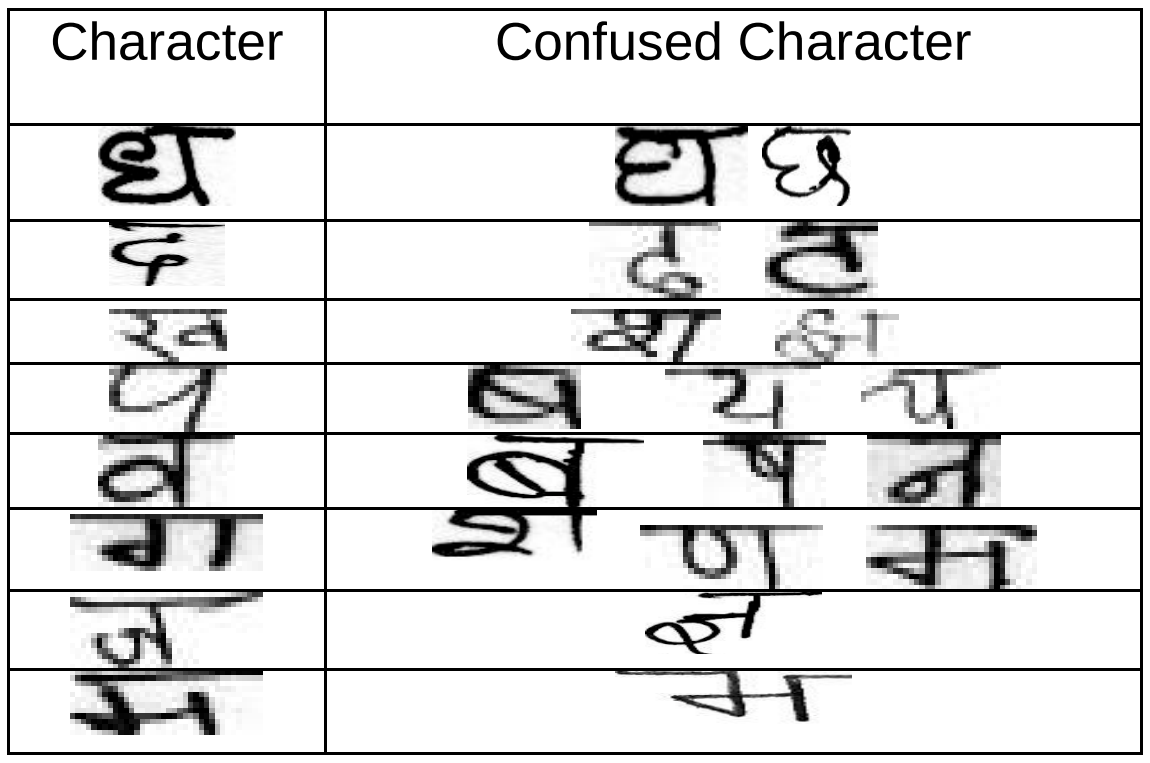

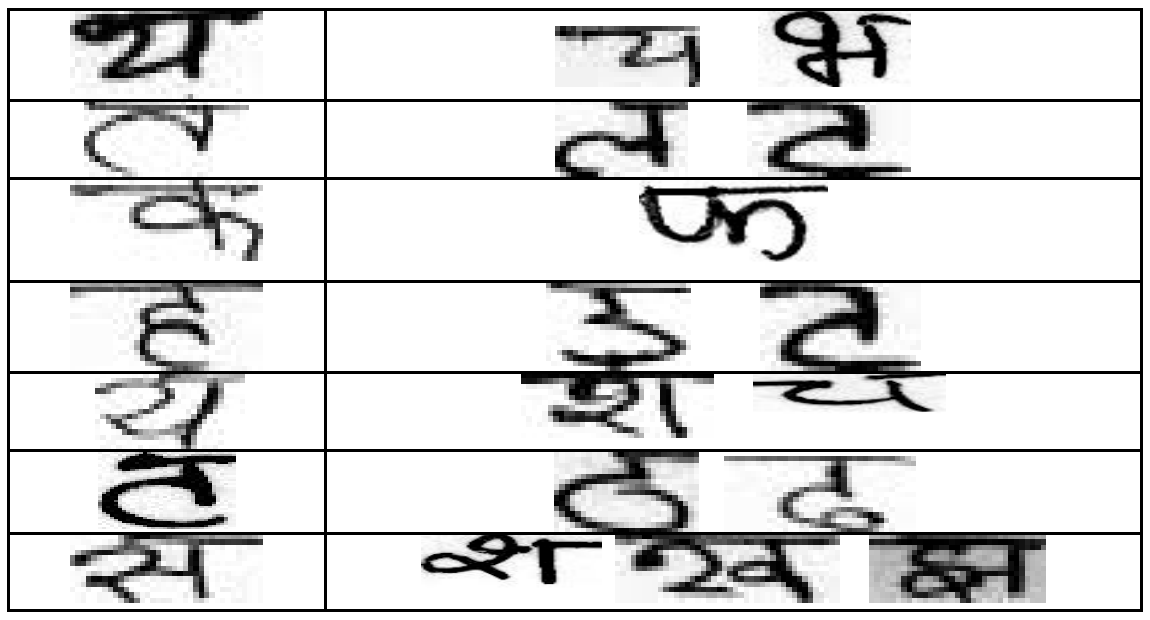

**TABLE 2** . Confused Characters pair

### 6.4 Comparison of results

We compared our current results with those existing pieces of work. Details comparative results are given in Table 3. In our previous work [20] we obtained 69.37% accuracy as top 1 choice result. Using this proposed technique we obtained 90.74% accuracy as top 1 choice result

| S. no. | Method purposed by | Data Size | Accuracy Obtained |
|---|---|---|---|
| 1. | Kumar and Singh [2] | 200 | 80% |
| 2. | N. Sharma, U. Pal, F. Kimura, and S. Pal [5] | 11270 | 80.36% |
| 3. | M. Hanmandlu, O.V. R. Murthy, V.K. Madasu[21] | 4750 | 90.65% |
| 4. | Proposed method | 7154 | 90.74% |

**TABLE 3.** Comparison of Results

## 7. CONCLUSION

The MLP and method of Minimum Edit Distance is tested on offline Handwritten Devnagari characters. MLP classifier is used on character set with certainty and minimum edit distance classifier is used on corner detected character image to classify the confused characters of similar shapes. Modified form of Harris corner detection algorithm is used for detecting corner in character image. Results are quite promising. In future we plan to experiment on other feature extraction methods and other classifiers to get higher recognition accuracy from our system.